\documentclass[runningheads]{llncs}

\usepackage{accv}

\usepackage{accvabbrv}

\usepackage{graphicx}
\usepackage{booktabs}
\usepackage{multirow}

\usepackage[accsupp]{axessibility}  

\usepackage{hyperref}

\usepackage{orcidlink}

\begin{document}

\title{A Recipe for CAC: Mosaic-based Generalized Loss for Improved Class-Agnostic Counting} 

\titlerunning{A Recipe for CAC}

\author{Tsung-Han Chou\inst{1}\orcidlink{0009-0009-7374-7221} \and
Brian Wang\inst{2}\orcidlink{0009-0004-6485-9217} \and
Wei-Chen Chiu\inst{1}\orcidlink{0000-0001-7715-8306} \and 
Jun-Cheng Chen\inst{3}\orcidlink{0000-0002-0209-8932}}

\authorrunning{TH Chou et al.}

\institute{National Yang Ming Chiao Tung University, Taiwan \and
University of Michigan, Ann Arbor \and
Research Center of Information Technology Innovation, Academia Sinica, Taiwan}

\maketitle

\begin{figure}[h]
\centering
\includegraphics[width=\linewidth]{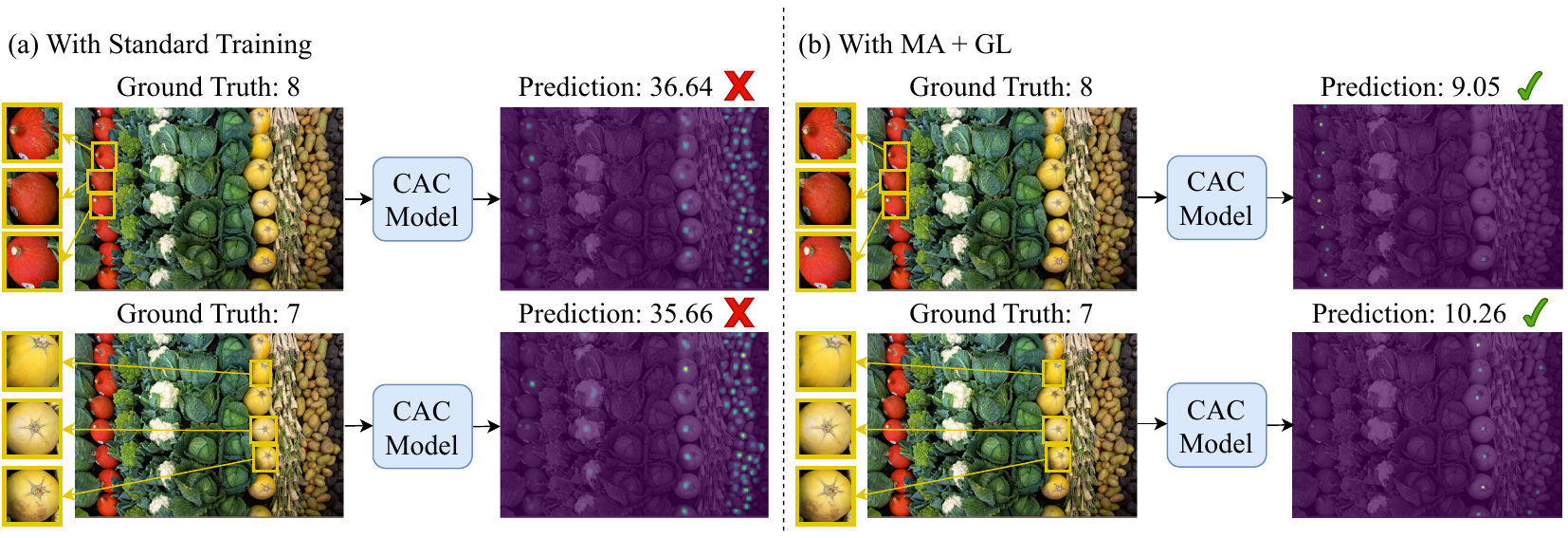}
\caption{\textbf{Comparison of CAC Models with Different Training Strategies in Real-World Images.} Given the same query image with different references (annotated by yellow bounding boxes), (a) the state-of-the-art CAC models (e.g., BMNet~\cite{shi2022represent}, LOCA~\cite{djukic2022low}) adopting
the standard training setting upon the FSC-147 dataset fail to distinguish the appearance of objects and gives wrong counts (i.e. the model does not fire at the objects of the same class as references on the query image, but the objects of majority. Noting that the dilation is applied on the density maps for best viewing). Conversely, (b) our suggested recipe, which utilizes novel technique of integrating Mosaic Augmentation (MA) and Generalized Loss (GL), enables the models to effectively discriminate different objects, thereby resulting in significantly better count prediction.}
\label{fig:comparasion_teasor}
\end{figure}

\begin{abstract}
Class agnostic counting (CAC) is a vision task that can be used to count the total occurrence number of any given reference objects in the query image. 
The task is usually formulated as a density map estimation problem through similarity computation among a few image samples of the reference object and the query image. In this paper, we point out a severe issue of the existing CAC framework: Given a multi-class setting, models don't consider reference images and instead blindly match all dominant objects in the query image. Moreover, the current evaluation metrics and dataset cannot be used to faithfully assess the model's generalization performance and robustness. To this end, we discover that the combination of mosaic augmentation with generalized loss is essential for addressing the aforementioned issue of CAC models to count objects of majority (i.e. dominant objects) regardless of the references. Furthermore, we introduce a new evaluation protocol and metrics for resolving the problem behind the existing CAC evaluation scheme and better benchmarking CAC models in a more fair manner.
Besides, extensive evaluation results demonstrate that our proposed recipe can consistently improve the performance of different CAC models. 
The code is available at \href{https://github.com/littlepenguin89106/MGCAC}{https://github.com/littlepenguin89106/MGCAC}. 
\end{abstract}    
\begin{figure}[t]
\centering
\includegraphics[width=\linewidth]{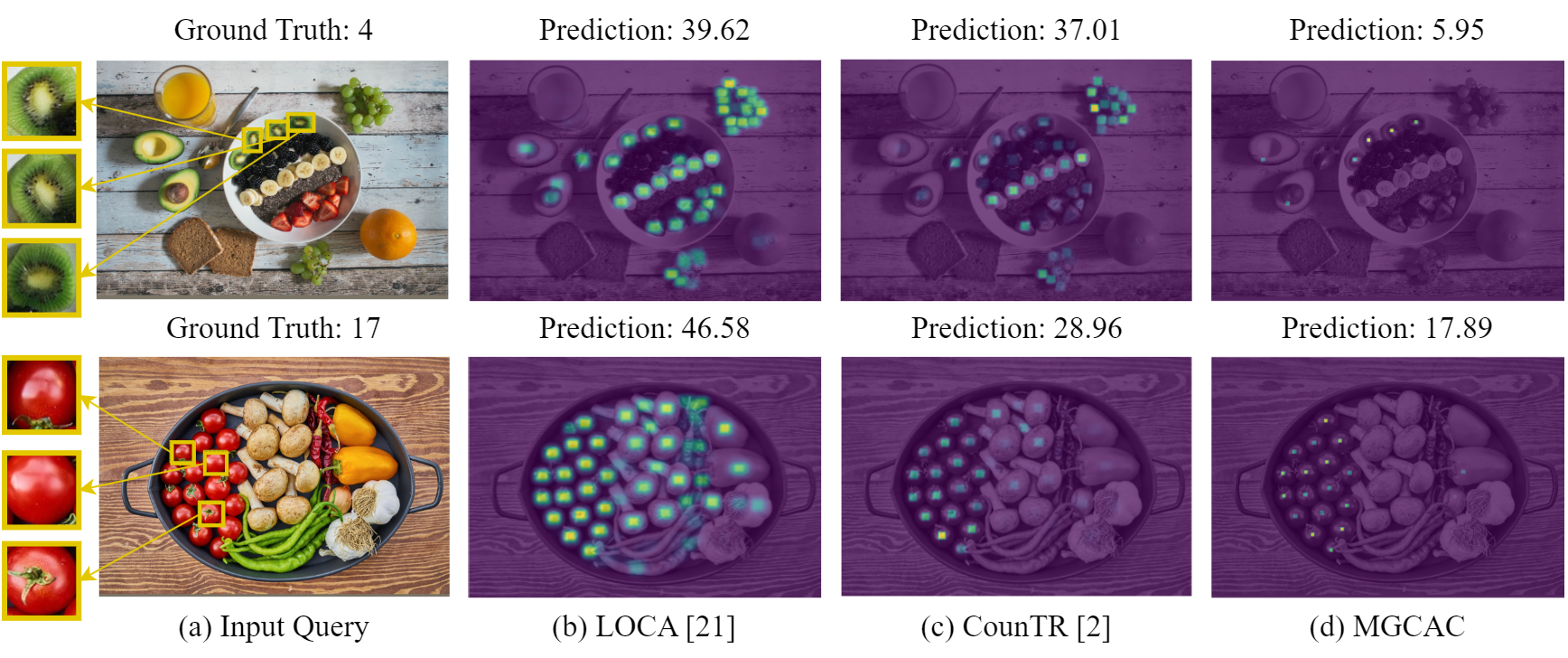}
\caption{\textbf{More qualitative results of various CAC models upon real images.} The existing state-of-the-art CAC models such as (b) LOCA~\cite{djukic2022low} and (c) CountTR~\cite{liu2022countr} have problems on either counting all objects in the query image (and completely disregarding the target objects showcased in reference images) or unintentionally detecting irrelevant objects. In contrast, our proposed MGCAC model, featuring the multi-class training scenario and the more feasible objective to CAC task, faithfully follows the guidance from the references to achieve superior results of counting.}
\label{fig:real_world}
\end{figure}
\section{Introduction}
\label{sec:intro}
Object counting is a popular research topic in the vision community with a wide spread of applications, including visual surveillance, intelligent agriculture, etc. It aims to count the occurrence number of target objects in an image. Object counting methods can be classified into two major categories: class-specific object counting and class-agnostic counting. Class-specific object counting usually focuses on counting a specific category such as cars, animals, people, etc, whereas among these applications, crowd counting, which aims to count the number of people in an image, is well studied by \cite{song2021rethinking,cheng2022rethinking,cheng2019learning,li2018csrnet}. However, class-specific counting requires training an individual model for each category with tremendous efforts on collecting thousands of training images with annotations and fails to work for unseen classes.\\
\indent In contrast, class-agnostic counting (CAC) studied by \cite{lu2018class, ranjan2021learning, yang2021class} arises recently and aims to count any novel objects within the query image, especially for those objects of unseen classes during the training stage.
Ideally, CAC should consider the appearance of $K$ reference images and count occurrences of similar objects in the query image. Nevertheless, in the existing evaluation protocol, query images typically only contain objects from a single class (even with clean backgrounds), which significantly differ from real-world multi-class cases. By our examination of real-world examples, we observed that most prior works degenerate back to blindly capture objects without considering the references (Figure \ref{fig:comparasion_teasor}(a) and Figure~\ref{fig:real_world}). We term the problem as Reference Oversight (RO). To verify RO, we propose a novel multi-class mosaic evaluation dataset for CAC. \\
\indent Secondly, we disclose that current benchmarking protocols are not fair indications of model generalization. Given FSC-147's greatly skewed per-image-object-count distribution (noting that FSC-147~\cite{ranjan2021learning} is a popular and represenatative dataset of CAC), we show that evaluation metrics such as Mean Absolute Error (MAE) and Root Mean Square Error (RMSE) tend to strongly encourage models to only optimize counting on images with high object counts while overlooking the remaining images. To ensure fair and comprehensive comparisons, we also incorporate metrics such as Normalized Absolute Error (NAE) and Squared Relative Error (SRE).\\
\indent Through in-depth analysis, we pinpoint the root causes of RO problem: the lack of variety of classes in each query image during the training stage incapacitates the model's need to differentiate among object classes. Additionally, using pixel-wise losses (e.g., MSE) fails to accurately localize target objects and still contributes to the model's inability to differentiate objects effectively. To address these challenges, we propose the Mosaic-based Generalized-Loss (MG) recipe for CAC, which consistently allows different CAC models to effectively capture variations in reference appearances and precisely localize target objects as illustrated in Figure \ref{fig:comparasion_teasor}(b) and demonstrated by our extensive experimental results. Finally, we also provide a strong CAC baseline by exploiting more advanced cross-attention modules from~\cite{cui2022mixformer} as the backbone with our proposed MG loss recipe for training, denoted as MGCAC. The proposed MGCAC achieves the state-of-the-art results for CAC upon the original FSC-147~\cite{ranjan2021learning} and our proposed mosaic evaluation dataset.\\
\indent Our contributions are summarized as follows: Firstly, our research is pioneering in exposing the performance gap of CAC methods between the existing evaluation protocol and real-world, multi-class scenarios. Our proposed mosaic evaluation dataset emphasizes RO problem on all existing CAC models, even the state-of-the-art methods (e.g. LOCA \cite{djukic2022low} or CounTR \cite{liu2022countr}) have such critical flaws (Figure~\ref{fig:real_world}(b) and \ref{fig:real_world}(c)). Secondly, to improve the robustness and localization of CAC, we introduce MG, a recipe that proves generalizable to various CAC models, and build a strong baseline, MGCAC. Our work underscores the importance of effective detection and discrimination of multi-class objects and hopes to reveal the ``devil'' hidden under the details of the current CAC framework.
\section{Related Works}\label{sec:related}

\subsection{Class-specific Object Counting}
Class-specific counting focuses on the task of counting the occurrence number of the objects of specific classes, such as crowd counting. In addition, the task is usually approached in two paradigms: detection-based and regression-based methods. The detection-based methods, including the methods proposed by \cite{leibe2005pedestrian,hsieh2017drone}, perform explicit object detection over the input image using visual object detectors and then get the count. However, similar to the object detection methods, their performances are also sensitive to situations when the objects are overlapped, occluded, or crowded. To address these problems, the regression methods proposed by \cite{thanasutives2021encoder,ma2021towards,cheng2022rethinking} instead predict the density map of input images where each pixel value can be interpreted as the fraction or the confidence level of the target object present in the query image. The sum of these values is then used as the estimated object count. In addition, the ground truth density maps for training are generated by convolving point annotations of the training images with properly selected Gaussian kernels. Although class-specific counting methods have achieved satisfactory results, they usually require training an individual network for each class of objects and can only be applied for counting the classes that have been seen in the training data.

\subsection{Class-Agnostic Counting}
Concerning class-specific object counting, class-agnostic counting aims to perform object counting for every class. 
In its setting, the models usually take a query image and several reference images from the same class as inputs and predict the count of the class that appears in the query image through the density map regression. Lu et al.~\cite{lu2018class} propose GMNet which is the first CAC framework, where the reference and query feature maps are extracted independently through the feature extractor based on the ResNet~\cite{he2016deep}. These features are then directly exploited as a pixel-wise regression task to perform object counting. To better capture the interactions between the query and the references during the matching stage, Ranjan et al.~\cite{ranjan2021learning} propose FamNet which introduces template matching for the CAC task. FamNet enhances the matching framework by convolving the reference feature maps across query feature maps. Each pixel value in the resulting query feature map represents the similarity between the query image and reference images at that specified location and therefore supports better localization. Building upon the existing structures, CFOCNet proposed by Yang et al.~\cite{yang2021class} matches query and references feature maps from different stages of backbone for multi-scale strategy. Moreover, CFOCNet integrates self-attention to strengthen the distinct features of the query image. Shi et al.~\cite{shi2022represent} propose BMNet which adopts bilinear similarity for matching. The bilinear similarity is a special case of generalized inner product and provides more flexibility for matching than previous methods using convolution. SPDCN\cite{lin2022glcac} introduces Scale-Prior Deformable Convolution and Scale-Sensitive Generalized Loss to enhance the model scale-awareness. CounTR proposed in~\cite{liu2022countr} introduces transformer-based architecture into class-agnostic counting, which utilizes attention to capture the similarity between the query and the references. Djukic et al. \cite{djukic2022low} propose a low-shot object counting network with iterative prototype adaptive (LOCA) which introduces the object prototype extraction module (OPE). The OPE module focuses on constructing strong object prototypes to improve object localization accuracy and iteratively adapts shape information and object appearances into object prototypes by recursively performing cross-attention operations. 

With the progress of multi-modality, recent works \cite{AminiNaieni23} and \cite{clipcount} adopt CLIP \cite{Radford2021LearningTV} model to perform CAC with input prompts. Given a class description and a query, by matching the extracted text features and the query features, the objects that match the description in the query will be identified. 

\section{Deficiencies of Current CAC Training and Evaluation}
In this section, we first introduce the general CAC framework. 
After that, we outline the blindly counting problem and the biases of the prevalent benchmark.
We provide our solutions to resolve the problem in Section~\ref{sec:method2} and show their effectiveness in Section~\ref{sec:exp}.
\begin{figure}[t]
\centering
\includegraphics[width=0.8\linewidth]{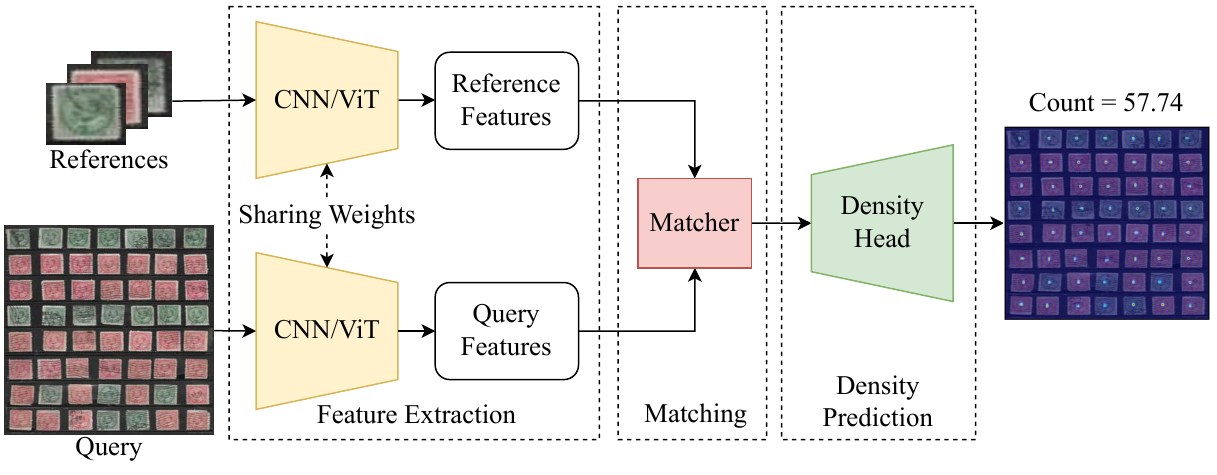}
\caption{\textbf{CAC Framework.} The general pipeline of CAC models consists of a feature extractor,
matcher, and density head. Given a query image and $K$ reference images (i.e., $K$ = 3 in this diagram), the model learns to predict a density
map for counting.}
\label{fig:cac_framework}
\end{figure}
\subsection{CAC Framework}\label{sec:background_cac}
As shown in Figure~\ref{fig:cac_framework}, standard CAC models comprise three components: (1) a feature extractor for visual feature extraction from reference and query images, (2) a cross-matching module for similarity map computation, and (3) a density map estimator for counting. Given a query image $X \in \mathbb{R}^{H_X \times W_X \times 3}$, $K$ reference images $Z=\{z_i\}_{i=1}^K, z_i \in \mathbb{R}^{H_{Z_i} \times W_{Z_i} \times 3}$, and a ground truth label $Y \in \mathbb{R}^{H_X \times W_X}$ (in the form of density map), the CAC model minimizes the regression problem to predict the number of the references occurring in the query image:
\begin{equation}
\min_{\theta, \phi, \psi} \mathbb{E}_{(X,Z,Y)}[\mathcal{L}( R_{\theta}(S_{\phi}(F_{\psi}(X), F_{\psi}(Z)), Y)],
\end{equation}
where $F_{\psi}(\cdot)$ is a pre-trained feature extractor, $S_{\phi}(\cdot,\cdot)$ computes the similarity map, $R_{\theta}(\cdot)$ is the regressor or density map estimator, and $\mathcal{L}(\cdot, \cdot)$ is the loss function (e.g., pixel-wise L2 loss). \\
\subsection{Reference Oversight}\label{sec:correct}
A robust CAC model should accurately count and localize objects corresponding to reference images in diverse query scenarios. However, most models are evaluated on widely used CAC datasets (e.g., FSC-147 \cite{shi2022represent} and CARPK \cite{hsieh2017drone}), which primarily feature a single-class. Such datasets lack real-life variability in object distributions and backgrounds. This setting raises concerns about the model's ability to precisely match target objects in multi-class scenarios.
As shown in Figure~\ref{fig:real_world}, we observe that most CAC models, including the current state-of-the-art (SOTA) LOCA, tend to annotate all dominant objects in query images, neglecting the similarity between objects in query and reference images. 
In contrast, CounTR leverages mosaic augmentation and therefore better identifies target objects. However, CounTR fails to calibrate the correct count for reference images due to inadequate localization information.
With regards to supervised CAC models, we hypothesize that: (1) The dominant issues are caused by training with queries dominated by objects of a single class, which CounTR uses mosaic augmentation that alleviates the issue (2) Supervision with pixel-wise L2 loss is not sensitive to localization on the object-level and leads to unsatisfied performance for CounTR.
\subsection{Benchmarking}\label{sec:benchmark}
A well-designed evaluation protocol should encompass diverse scenarios and utilize fair metrics. However, existing CAC datasets, as discussed in Section \ref{sec:correct}, often lack diversity of categories within a single query. Conversely, detection datasets like COCO \cite{lin2014microsoft} include multiple classes but predominantly have a small count of objects per image, failing to comprehensively assess the model's ability to identify corresponding objects in queries. 

\begin{figure}[ht]
    \begin{minipage}[b]{0.49\textwidth}
    \centering
    \includegraphics[width=\textwidth]{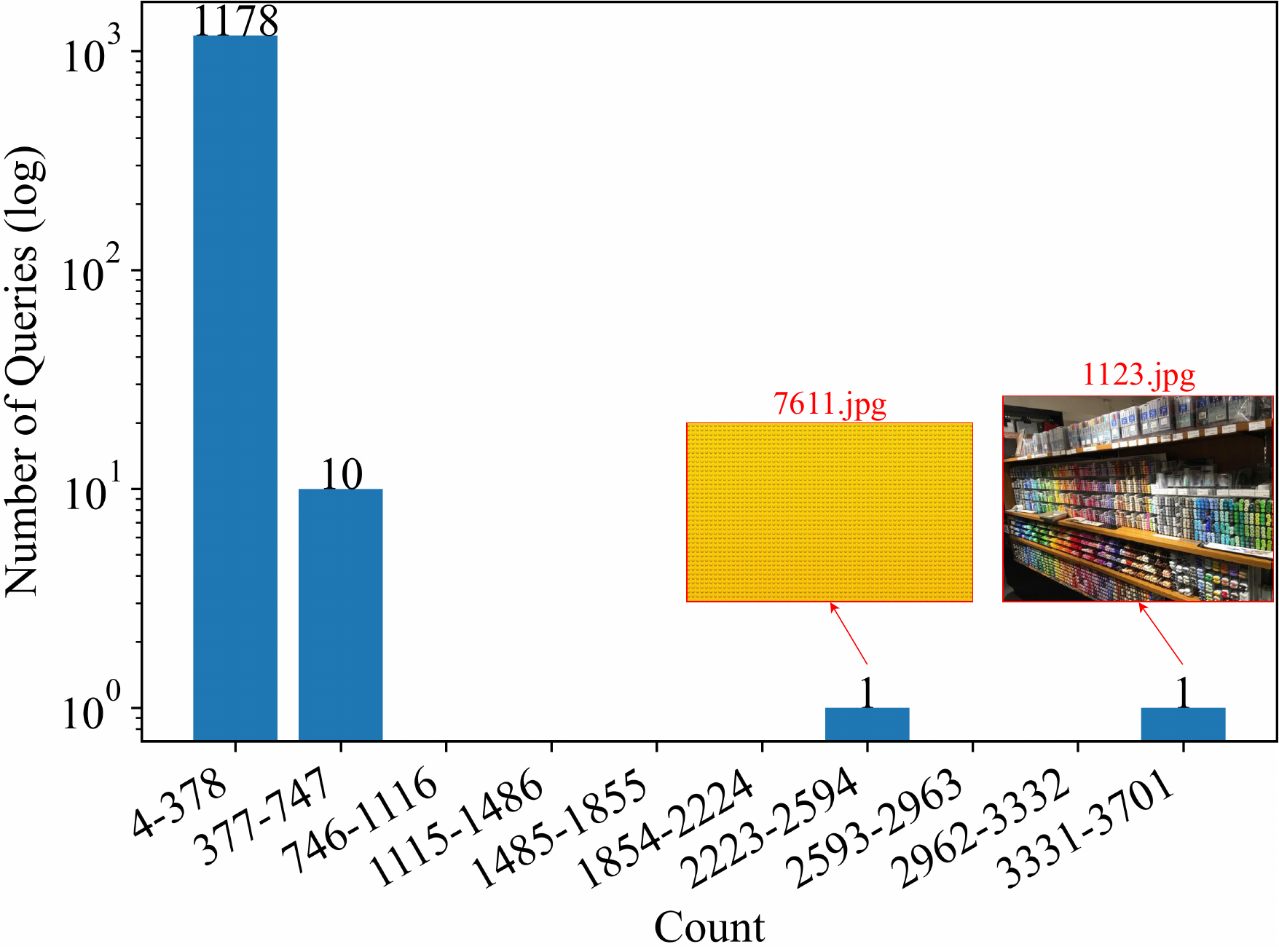}
    \captionof{figure}{\textbf{Distribution of object count in FSC-147 test dataset.} We split the test dataset into 10 bins. The x-axis is the count range of the corresponding bin, y-axis is the number of queries in log-scale.}
    \label{fig:bar}
    \end{minipage}
    \hfill
    \begin{minipage}[b]{0.49\linewidth}
        \centering
        \includegraphics[width=\textwidth]{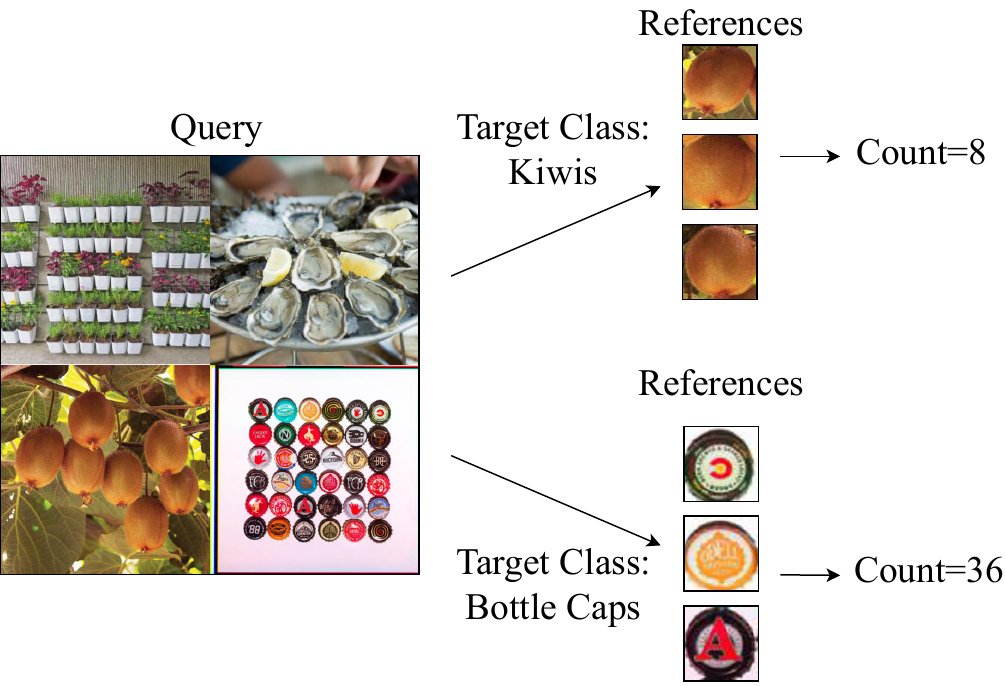}
        \captionof{figure}{\textbf{Mosaic evaluation.} Given different references of the target class, the objective is to count the corresponding objects in the query.}
        \label{fig:mosaic_pipeline}
    \end{minipage}
\end{figure}

Furthermore, the current CAC evaluation protocol evaluates model counting performance by aggregating counting errors across all images. Given groundtruth count $c_l$ and predicted count $\tilde{c_l}$ for the $l$th image in the dataset with $L$ images, \(\text{MAE} = \frac{1}{L}\sum_{i=1}^L |\tilde{c_l} - c_l|\) and \(\text{RMSE} = \sqrt{\frac{1}{L}\sum_{i=1}^L (\tilde{c_l} - c_l)^2}\) are usually used.
The challenge arises from the long-tailed distribution of per-image object counts, introducing bias towards large-count query images. Figure \ref{fig:bar} illustrates that only two images featuring large counts in FSC-147 test dataset.
By excluding the first two highest object-count query images from RMSE calculations on the FSC-147 test dataset (Table \ref{tab:exclude}), we show significant impacts on the evaluation results (details in Appendix). Recently SOTA models (i.e., LOCA and CounTR) get worse RMSE than BMNet+ when removing large-count queries.
Those phenomena reinforce the need for an evaluation strategy that accounts for the distributional characteristics of object counts, especially in datasets with varying count sizes.
\begin{table}[t]
\caption{\textbf{Comparison of the performance of CAC models on the full FSC-147 test dataset and the dataset without outliers (i.e., 1123.jpg and 7611.jpg).}}
\begin{subtable}{0.49\textwidth}
\centering
\caption{Full test set.}
\begin{tabular}{ccccc} \toprule
       & MAE   & RMSE  & NAE  & SRE  \\ \hline
BMNet+ \cite{shi2022represent} & 14.62 & 91.83 & 0.27 & 6.20 \\
CounTR \cite{liu2022countr} & 11.95 & 91.23 & 0.23 & 7.44 \\
LOCA \cite{djukic2022low}  & 10.79 & 56.97 & 0.19 & 2.19 \\ \bottomrule
\end{tabular}
\end{subtable}
\begin{subtable}{0.49\textwidth} 
\centering
\caption{Test set after excluding outliers}
\begin{tabular}{ccccc} \toprule
       & MAE   & RMSE  & NAE  & SRE  \\ \hline
BMNet+ \cite{shi2022represent} & 12.04 & 30.13 & 0.27 & 6.04 \\
CounTR \cite{liu2022countr} & 9.93  & 33.61 & 0.23 & 6.47 \\
LOCA \cite{djukic2022low} & 9.46  & 33.03 & 0.18 & 1.94 \\ \bottomrule
\end{tabular}
\end{subtable}
\label{tab:exclude}
\end{table}
\section{Proposed Evaluation Protocol}
\subsection{Mosaic Evaluation Dataset}
Currently, there is a lack of real-world multi-class counting datasets for CAC evaluation. We leverage the FSC-147 dataset as a foundation to create a multi-class setting. As shown in Figure~\ref{fig:mosaic_pipeline}, for a synthesized mosaic collage, we randomly select 4 query images of distinct classes from the validation and test sets of FSC-147. These images are then randomly cropped to the size of 384 × 384 and put together as size 768 × 768. Each class in the collage will individually serve as the reference class. Given the associated reference images, the model has to accurately predict the number of target objects for the collage (i.e. predicted count should equate to the ground truth count of the query image containing the target class).
In total, we generate 1,000 queries (i.e., 4,000 data pairs), facilitating a comprehensive evaluation of the object recognition capabilities within the context of mosaic collages. We name this dataset as FSC-Mosaic, which stands for
\underline{F}ew-\underline{S}hot \underline{C}ounting Dataset with \underline{Mosaic} Evaluation.
\subsection{Additional Metric}
As discussed in Section \ref{sec:benchmark}, using MAE and RMSE is heavily influenced by query images with large object counts. For balancing the contribution of each image to the performance score, we additionally benchmark with \(\text{NAE} = \frac{1}{L}\sum_{l=1}^L \frac{|\tilde{c_l} - c_l|}{c_l}\) and \(\text{SRE} = \sqrt{\frac{1}{L}\sum_{i=1}^L \frac{(\tilde{c_l} - c_l)^2}{c_l}}\), which provide a better reflection of the model's overall performance on the entire dataset. As shown in Table \ref{tab:exclude}, the use of NAE and SRE provides a more stable assessment of model performance on datasets with skewed object count distributions.
\section{Proposed MG Recipe for CAC Training}\label{sec:method2}
To improve the class discrimination of CAC models and their matching performance on targets, we tackle it in two ways: (1) exposing multi-class setting through mosaic augmentation, and (2) leveraging Generalized Loss (GL) that is more localization-aware and robust. We further elucidate how the combination of these strategies improves the RO problem.
\subsection{Mosaic Augmentation}\label{sec:mosaic_aug}
The main issue for the CAC domain is its single-class per-image data setting, which hinders the model's generalizability under multi-class setting. 
To directly transform single-class setting to multi-class setting, we apply mosaic augmentation (MA) on FSC-147 images. We construct it by (1) randomly selecting 4 images, each randomly cropping to dimensions $H_Z \times W_Z$, (2) arranging it into $2H_Z \times 2W_Z$ mosaiced image. During training, we randomly select 1 out of the $M$ reference classes in the mosaic for prediction. As compared to the original single-class image, the model is now exposed to $M$ object classes at once, therefore acquiring the capability to distinguish among different class objects while recognizing reference objects that slightly vary in appearance. Our experiments show the non-trivial impact of this data augmentation technique that greatly strengthens the model's robustness on both our mosaic evaluation protocol and real-life multi-class scenarios.
\subsection{Generalized Loss}
Currently, most CAC approaches based on density map estimation employ the pixel-wise $L_2$ loss (or MSE loss) to train the models where the ground truth density map is generated by converting the point annotations of each training image through convolving with a Gaussian kernel. 
However, $L_2$ loss only offers pixel-wise supervision. It penalizes the same no matter whether the predictions are near or far from the ground-truth dot annotation. This makes $L_2$ loss inappropriate for accurate localization.
Recently, Wan et al.~\cite{wan2021gl} propose a Generalized Loss (GL) that uses unbalanced optimal transport (OT) to measure the transport cost between the predicted density map and the ground truth point annotations directly. GL penalizes more when the predictions are farther from the ground truth. This suffices the need for precise localization for single and multi-class settings.
In addition, it also has shown that the $L_2$ loss is a special case of GL and usually results in suboptimal solutions. Thus, we mainly exploit GL to train the proposed models shown as follows:
\begin{equation}
L^{\tau}_C = \min_
{P} \langle C, P \rangle - \varepsilon H(P)
+ \tau \| P\mathbf{1}_m - a\|^2_2 + \tau \| P^{\top}\mathbf{1}_n - b\|_1,
\end{equation}
where $C$ is the transport cost of moving predicted density to ground truth dot annotation, $P$ is the corresponding transport plan, $H(\cdot)$ is the entropic regularization term, $n$ is the number of pixels, $m$ is the number of annotation points, $a$ is the predicted density map, and $b$ is the ground truth dot map.
Furthermore, to better encode perspective information, it proposes the Perspective-Guided Transport Cost:
\begin{equation}
C_{ij} = \exp(\frac{1}{\eta(x_i,y_i)}\|x_i-y_j\|_{2}),
\end{equation}
where $\eta(x_i,y_i)$ is a adaptive perspective factor. We simply choose a fixed $\eta$ for our experiment.
\subsection{Combination of MA and GL}
While both MA and GL are not new in counting tasks, previous studies have not fully exploited the potential of MA and GL in multi-class scenarios. For instance, CounTR utilizes MA for
resolving long-tailed problems (i.e. few images with large object counts), which differs from our intention of creating multi-class setting. Similarly, SPDCN uses GL for better localization in single-class setting, which can not distinguish different classes. By combining MA and GL, models have stronger awareness of reference locations, allowing them to discern the appearance of reference objects in multi-class settings. Furthermore, they also generalize well to count unseen real-world multi-class images as illustrated in Figure \ref{fig:comparasion_teasor}(b). In the following experiment, we show the necessity of our approach in bridging the domain gap between single and multi-class settings.
\section{Experiment}\label{sec:exp}
In this section, we present the details of the experimental setup. Then, we show extensive evaluation results and ablation studies of the proposed training recipe on the widely used CAC benchmarks. We further evaluate CAC models on our proposed evaluation protocol to examine CAC models' ability to localize reference targets under multi-class scenarios.
\subsection{Implementation Details} \label{sec:implementation}
\textbf{Datasets.} We utilize three datasets to evaluate our model performance: (1) FSC-147 dataset~\cite{ranjan2021learning}, a standard class-agnostic object counting dataset composed of 147 categories and 6,135 images across different scales, in which each image is annotated with three reference objects. Given that the dataset contains decent annotations, we use the FSC-147 dataset to train and evaluate our models. (2) We evaluate FSC-147-trained models on our proposed FSC-Mosaic dataset to validate their generalizability in multi-class scenarios. (3) Moreover, we evaluate FSC-147-trained models on CARPK~\cite{hsieh2017drone}, a dataset that contains 1,448 photos of cars in parking lots from a bird-view perspective, to ensure the generalizations of our models on unseen datasets (see Appendix). \\ 
\indent \textbf{Evaluation Metrics.} To compare performance against existing CAC models, we preserve the previous evaluation protocol~\cite{ranjan2021learning} based on MAE and RMSE. We additionally benchmark with NAE and SRE to ensure fair consideration of each image in the evaluation process.\\
\indent \textbf{Model Comparison.} We utilize MixFormer \cite{cui2022mixformer} architecture design with a multi-scale density head to better localize features of different levels as our baseline, denoted as MGCAC as mentioned in Section~\ref{sec:intro}. (Please refer to Appendix for more detailed descriptions.) It is worth noting that our training recipe can be applied to other CAC model architectures as well. To demonstrate the effectiveness and general applicability of our proposed training recipe, we compare BMNet+, CounTR, and LOCA with their original training approach against our proposed recipe on the FSC-Mosaic dataset. \\
\indent\textbf{Data Preprocessing.} 
For our baseline training preprocessing, query images are (1) randomly flipped horizontally, and (2) randomly cropped to the same size of $384 \times 384$. When applying mosaic augmentation, 4 images with different classes are randomly cropped to $192 \times 192$, so the synthesized mosaic collage keeps the same image size in a batch for efficient training. We randomly select 1 out of the 4 reference classes as the target, using the corresponding references and synthetic mosaic collage as input, with the corresponding annotations as ground truth. During the evaluation stage, the processing of reference images is similar to training, but query images are fully used. \\
\indent \textbf{Optimizer and Training Hyperparameters.} Our model is trained with 300 epochs, batch size of 8, optimized with AdamW optimizer, and weight decay of 5e-4. For MGCAC, the backbone learning rate is 1e-5, and the non-backbone learning rate is 1e-4. We use PyTorch~\cite{paszke2019pytorch} as the implementation framework. 
We use GL to supervise the final output that is $\frac{1}{4}$ of the original query image size.
The hyperparameters of GL are set as $\varepsilon = 0.01$, $\tau = 0.5$, and $\eta = 0.6$.\\
\indent \textbf{Inference time.} The running time of our MGCAC takes 38ms for one evaluation on average using a single NVidia RTX 3090 and an Intel i7-8700 CPU.
\subsection{Evaluation Results with Other SOTAs}
\begin{figure}[t]
     \centering
     \includegraphics[width=\textwidth]{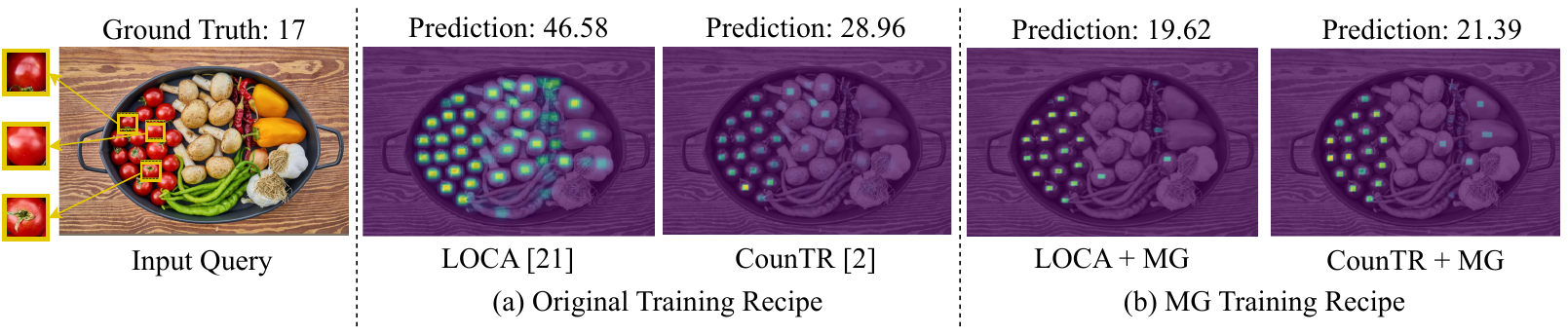}
    \caption{\textbf{Comparison of CAC models with original training strategy (official pre-trained models) and our proposed MG training recipe.}}
\label{fig:mg_recipe}
\end{figure}

\textbf{Evaluation on Existing FSC-147 Setting.}
As shown in Table~\ref{table:overall_results}, our model surpasses most of models on NAE and SRE, which are metrics that indicate counting errors across images, unweighted by object counts. As compared to the SOTA model LOCA, our performance yields an improvement of 45.4\% on VAL NAE, 1.9\% on VAL SRE, and 15.7\% on TEST NAE. These results indicate that our model performs better on average across each image. 
When considering the existing metrics MAE and RMSE, our MAE is comparable to LOCA's while having a gap in RMSE. As shown in Section \ref{sec:benchmark}, RMSE is not an indicative metric of model performance. We also demonstrate MGCAC* that intentionally overfits and optimizes the performance on a few large-count images by adopting test-time normalization from \cite{liu2022countr}, which acquires SOTA on the FSC-147 dataset. This validates the unbalanced contribution of RMSE in large-count images and our model superiority on non-outlier images. 
In the next subsection, we discuss the evaluation of multi-class settings, where the model's awareness of reference context is challenged. \\
\begin{table}[t]
\centering
\caption {Comparisons with SOTA CAC models on the FSC-147 dataset. Notation `$\ast$' indicates that we further adopt the test-time normalization~\cite{liu2022countr} during inference.}\label{table:overall_results}
\resizebox{0.8\columnwidth}{!}{%
\begin{tabular}{ccccccccc} \toprule
\multirow{2}{*}{Method} & \multicolumn{4}{c}{VAL (validation)}       & \multicolumn{4}{c}{TEST}       \\ 
                        & MAE$\downarrow$   & RMSE$\downarrow$  & NAE$\downarrow$   & SRE$\downarrow$   & MAE$\downarrow$   & RMSE$\downarrow$   & NAE$\downarrow$   & SRE$\downarrow$   \\ \hline
GMN \cite{lu2018class}                    & 29.66 & 89.81 & ----- & ----- & 25.52 & 124.57 & ----- & ----- \\
FamNet \cite{ranjan2021learning}                  & 23.75 & 69.07 & 0.51  & 4.24  & 22.08 & 99.54  & 0.44  & 6.45  \\
CFOCNet \cite{yang2021class}                & 21.19 & 61.41 & ----- & ----- & 22.10 & 112.71 & ----- & ----- \\
BMNet+ \cite{shi2022represent}                 & 15.74 & 58.53 & 0.25  & 2.73  & 14.62 & 91.83  & 0.27  & 6.20  \\
CounTR \cite{liu2022countr}                 & 13.13 & 49.83 & 0.23  & 2.59  & 11.95 & 91.23  & 0.23  & 7.44  \\
SAFECount \cite{you2023few}              & 15.28 & 48.20 & 0.26  & 3.99  & 14.32 & 85.54  & 0.25  & 6.37  \\
SPDCN \cite{lin2022glcac}                  & 14.59 & 49.97 & 0.22  & 2.79  & 13.51 & 96.80  & 0.22  & 6.70  \\
LOCA \cite{djukic2022low}                   & 10.24 & \textbf{32.56} & 0.22  & 2.09  & 10.79 & 56.97  & 0.19  & \textbf{2.19}  \\ \hline
\textbf{MGCAC (Ours)}                  & 11.00 & 51.42 & \textbf{0.12}  & 2.05  & 10.46 & 96.60  & \textbf{0.16}  & 6.17  \\
\textbf{MGCAC* (Ours)}                  & \textbf{9.93}  & 41.08 & \textbf{0.12}  & \textbf{1.91}  & \textbf{9.10}  & \textbf{54.21}  & \textbf{0.16}  & 5.68 \\ \bottomrule
\end{tabular}}
\end{table}
\indent \textbf{Mosaic Evaluation.}
Seen from Table~\ref{tab:mosaic_evaluation}, models under the CAC typical training process, such as BMNet+ and LOCA, exhibit unsatisfactory performance under mosaic evaluation. The introduction of mosaic augmentation, as demonstrated by CounTR, yields a better performance. After employing our training recipe, BMNet+, CounTR, and LOCA exhibit substantial error decreases in all metrics. Specifically, BMNet+ yields an improvement of 47.5\% in RMSE and over 75\% across the remaining three metrics. These improvements demonstrate the importance and efficacy of applying the combination of mosaic augmentation and generalized loss on CAC models in discriminating reference classes and accurately matching targets. Notably, MGCAC achieves SOTA on the FSC-Mosaic dataset with 37\% improvement on all metrics against all models, showing our superiority in counting target objects. \\
\indent \textbf{Real-world examples.}
To verify whether the CAC models are general for real-world scenarios, we test them with some real-world examples as well. As shown in Figure \ref{fig:real_world}(d), MGCAC excels in accurately localizing target objects and providing relatively precise prediction counts. Notably, as shown in Figure \ref{fig:mg_recipe}, LOCA and CounTR, trained with our MG recipe, efficiently locate target objects and mitigate the RO problem, validating the effectiveness and generability of our training recipe.\\
\begin{table}[t]
\centering
\caption{Evaluation results on the FSC-Mosaic dataset of the CAC models trained with the proposed training recipe (\textcolor{Red}{Red}) against those using their original training recipe where we use their publicly available pre-trained models (\textcolor{Green}{Green}).}
\label{tab:mosaic_evaluation}
\begin{tabular}{ccccccc}
\toprule
MA & GL & Model                                             & MAE                            & RMSE                            & NAE                            & SRE                            \\ \hline
 $\times$  &  $\times$  & \textcolor{Green}{FamNet} \cite{ranjan2021learning} & 37.09                          & 66.63                           & 1.65                           & 10.57                          \\
 $\times$  &  $\times$  & \textcolor{Green}{BMNet+} \cite{shi2022represent}   & 81.69                          & 101.75                          & 5.28                           & 24.73                          \\
 $\times$  &  $\times$  & \textcolor{Green}{LOCA} \cite{djukic2022low}        & 42.24                          & 67.19                           & 2.45                           & 13.33                          \\ \hline
 \checkmark & $\times$ & BMNet+ \cite{shi2022represent} & 24.38 & 52.81 & 1.17 & 7.81 \\
 \checkmark  &  $\times$  & \textcolor{Green}{CounTR} \cite{liu2022countr}      & 26.07                          & 65.04                           & 0.51                           & 4.55                           \\ \hline
  $\times$  &  \checkmark  & \textcolor{Green}{SPDCN} \cite{lin2022glcac}        & 40.60                          & 85.38                           & 0.99                           & 6.36                           \\ \hline
 \checkmark  &  \checkmark  & \textcolor{Red}{BMNet+} \cite{shi2022represent}   & 20.51                          & 53.37                           & 0.58                           & 5.65                           \\
 \checkmark  &  \checkmark  & \textcolor{Red}{CounTR} \cite{liu2022countr}      & 23.77                          & 63.44                           & 0.43                           & 4.40                           \\
 \checkmark  &  \checkmark  & \textcolor{Red}{LOCA} \cite{djukic2022low}        & 27.69                          & 64.36                           & 0.64                           & 4.66                           \\
 \checkmark  &  \checkmark  & \textcolor{Red}{\textbf{MGCAC (Ours)}}             & \textbf{8.58} & \textbf{33.28} & \textbf{0.17} & \textbf{1.93} \\ \bottomrule
\end{tabular}
\end{table}
\begin{figure}[htb]
     \centering
     \includegraphics[width=0.65\textwidth]{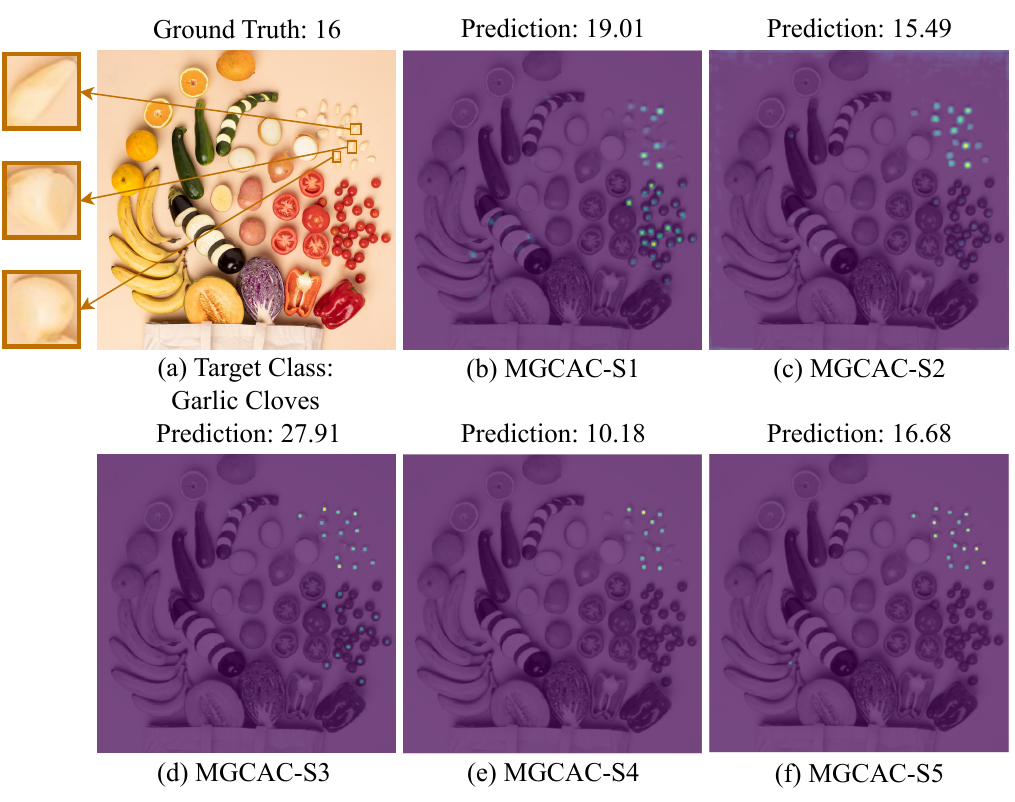}
    \caption{\textbf{Ablation of MGCAC (cf. Section~\ref{ablation})}. We visualize the predicted density map results of our MGCAC on the real image. The top-left corner is the input query image with references (annotated with bounding boxes), while its caption provides information about the number and the class of target objects.}
\label{fig:ablation}
\end{figure}
\subsection{Ablation Studies}\label{ablation}
We perform ablation studies on MGCAC to examine the functionality of our proposed framework in the FSC-147 dataset and FSC-Mosaic dataset. Based on the results, we have the following observations:\\
\indent \textbf{Mosaic Augmentation.} In Table~\ref{tab:ablation_fsc}, comparing S1 and S2, the introduction of mosaic augmentation (S2) worsens the performance on FSC-147 dataset. In contrast, S2 gets significant improvement on the FSC-Mosaic dataset (Table~\ref{tab:ablation_mosaic}). From Figure \ref{fig:ablation}(c), we observe that mosaic augmentation allows S2 to focus on the correct region. Nevertheless, S2 still assigns density values on irrelevant objects and gives inaccurate counts. \\
\indent \textbf{Generalized Loss.} With only applying GL (comparing S1 and S3), although S3 gets huge improvement on FSC-147 dataset (Table \ref{tab:ablation_fsc}), its performance becomes worse on FSC-Mosaic dataset (Table \ref{tab:ablation_mosaic}). As shown in Figure \ref{fig:ablation}(d), although S3 captures the target objects, it also includes irrelevant objects. These observations indicate that while the use of GL enhances the localization ability, it inadvertently leads to model overfitting in single-class settings, hindering its effectiveness in differentiating classes.\\
\indent \textbf{Combination of MA and GL.}
Comparing S2 and S4 quantitatively, the introduction of GL yields a relative improvement of $10\%$ in MAE, $26\%$ in NAE, and $7\%$ in SRE for our MGCAC. 
By combining MA and GL, the model has stronger supervision under multi-class setting, allowing them to discern the appearance of reference objects.
The large improvement in object localization is shown in Figure~\ref{fig:ablation}(e). While the count is not accurate, the crucial point is that S4 effectively captures only the target objects, excluding irrelevant ones. \\
\indent \textbf{Multi-scale Feature with U-Net-like Fusion.} Comparing S4 and S5 quantitatively, the incorporation of multi-scale feature aggregation (i.e. the multi-scale features from the feature extractor and the matcher are fused via a U-Net-like density head) allows MGCAC to be competitive on the FSC-147 dataset while still demonstrating its ability to distinguish reference classes on the FSC-Mosaic dataset. The predicted count closely aligns with the ground truth in Figure~\ref{fig:ablation}(f).\\

\begin{table}[t]
\centering
\caption{The ablation results of the proposed MGCAC on FSC-147 and FSC-Mosaic dataset, where `MA' denotes using mosaic augmentation, `GL' denotes replacing the MSE loss with the generalized loss, `MS' means that the multi-scale features from the feature extractor and the matcher are fused via a U-Net-like density head.
}
\begin{subtable}[h]{\linewidth}
\centering
\caption{Ablation results on FSC-147 dataset.}
\centering
\resizebox{0.95\columnwidth}{!}{%
\setlength{\tabcolsep}{5pt}
\begin{tabular}{cccccccccccc} \toprule
\multirow{2}{*}{Setting} & \multirow{2}{*}{MA} & \multirow{2}{*}{GL} & \multirow{2}{*}{MS} & \multicolumn{4}{c}{VAL} & \multicolumn{4}{c}{TEST} \\
   &  &  &  & MAE   & RMSE  & NAE  & SRE  & MAE   & RMSE   & NAE  & SRE  \\ \hline
S1 & $\times$ & $\times$ & $\times$ & 14.83 & 63.70 & 0.17 & 2.60 & 14.30 & 101.89 & 0.22 & 7.39 \\
S2 & \checkmark & $\times$ & $\times$ & 16.38 & 70.93 & 0.18 & 2.83 & 16.16 & 116.32 & 0.23 & 6.89 \\
S3 & $\times$ & \checkmark & $\times$ & 13.23 & 62.15 & 0.13 & 2.42 & 12.29 & 100.75 & 0.18 & 6.51 \\
S4 & \checkmark & \checkmark & $\times$ & 14.68 & 67.40 & 0.13 & 2.62 & 13.61 & 119.97 & 0.17 & \textbf{6.10} \\
S5 & \checkmark & \checkmark & \checkmark & \textbf{11.00} & \textbf{51.42} & \textbf{0.12} & \textbf{2.05} & \textbf{10.46} & \textbf{96.60}  & \textbf{0.16} & 6.17 \\ \bottomrule
\end{tabular}}
\label{tab:ablation_fsc}
\end{subtable}
\begin{subtable}[h]{\linewidth}
\centering
\caption{Ablation results on FSC-Mosaic dataset.}
\resizebox{0.65\columnwidth}{!}{%
\setlength{\tabcolsep}{5pt}
\begin{tabular}{clllcccc}
\toprule
\multicolumn{1}{l}{Setting} &
  MA &
  GL &
  MS &
  \multicolumn{1}{l}{MAE} &
  \multicolumn{1}{l}{RMSE} &
  \multicolumn{1}{l}{NAE} &
  \multicolumn{1}{l}{SRE} \\ \hline
S1 & $\times$ & $\times$ & $\times$ & 37.42 & 70.58 & 2.22 & 15.94 \\
S2 & \checkmark & $\times$ & $\times$ & 12.17 & 41.23 & 0.35 & 2.79  \\
S3 & $\times$ & \checkmark & $\times$ & 45.03 & 76.91 & 2.88 & 17.28  \\
S4 & \checkmark & \checkmark & $\times$ & 10.45 & 40.40 & 0.17 & 2.24  \\
S5 & \checkmark & \checkmark & \checkmark & \textbf{8.58} & \textbf{33.28} & \textbf{0.17} & \textbf{1.93} \\ \bottomrule
\end{tabular}}
\label{tab:ablation_mosaic}
\end{subtable}
\end{table}

\section{Conclusion}
In this work, we present a Mosaic-based Generalized Loss recipe that resolves the CAC model's failure to distinguish among reference classes. We analyze the skewed distribution phenomenon that hides under the optimization of existing metrics of MSE/RMSE. With our proposed evaluation protocol, we can effectively quantify the model's ability to learn discriminative reference features and accurately locate target objects in a multi-class setting. We also utilize different metrics (NAE, SRE) to represent the model's overall performance across images. Our mosaic augmentation and supervision of generalized loss have been empirically proven to significantly improve CAC models' robustness. With these contributions, we hope to set up a better standard for CAC domain that reflects model generalizability and robustness in real-life settings. 

\section{Acknowledgements}
This research work is supported in part by National Science and Technology Council (NSTC) under grants 112-2222-E-001-001-MY2, 112-2634-F-002-006, 112-2221-E-A49-087-MY3, 111-2628-E-A49-018-MY4, and by Academia Sinica under grant AS-CDA-110-M09.  We also thank to National Center for High-performance Computing (NCHC) of National Applied Research Laboratories (NARLabs) in Taiwan for providing computational and storage resources.

%
%
\bibliographystyle{splncs04}
\bibliography{references/cac,references/loss,references/tracking}
\end{document}


\title{Appendix} 

\titlerunning{A Recipe for CAC}

\author{}

\authorrunning{TH Chou et al.}

\institute{}

\maketitle

\begin{figure}[ht]
     \centering
     \includegraphics[width=\textwidth]{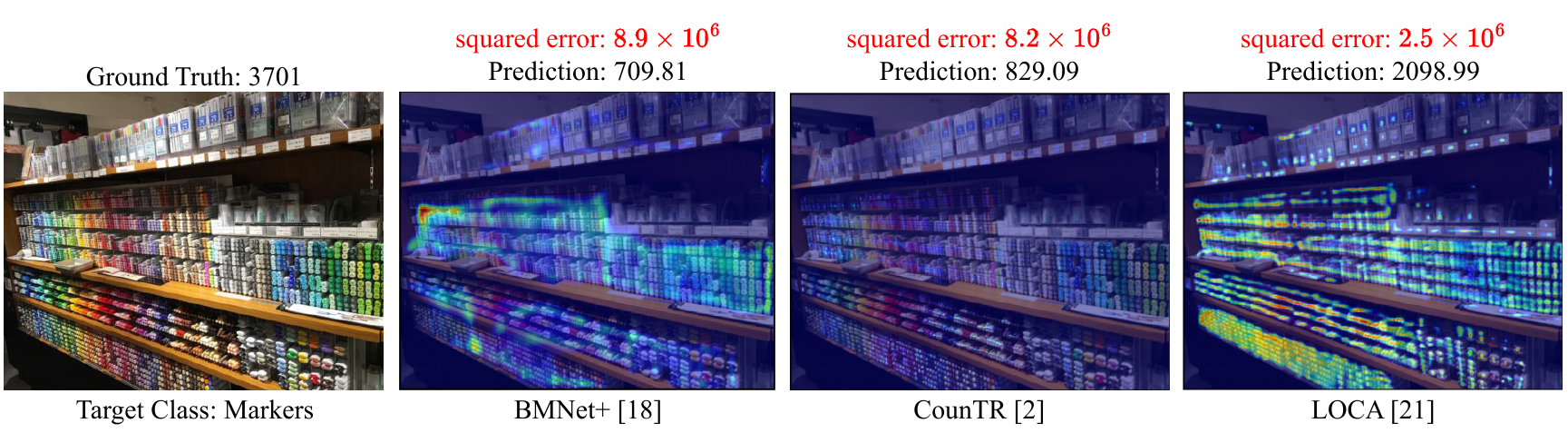}
    \caption{\textbf{Comparison of CAC models evaluated on 1123.jpg from FSC-147 dataset.} Both BMNet+ and CounTR exhibit a considerable difference in squared error compared to LOCA.}
\label{fig:1123}
\end{figure}

\section{Counting error of 1123.jpg on FSC-147 dataset}
In Section 3.3 of the main paper, we demonstrate that images with a high object count significantly impact MAE and RMSE scores. Specifically, we show that most CAC models produce large counting errors for 1123.jpg (Figure \ref{fig:1123}), which is an outlier in this dataset.

\section{Test-Time Normalization}
\begin{figure}[hb]
\centering
\includegraphics[width=0.6\textheight]{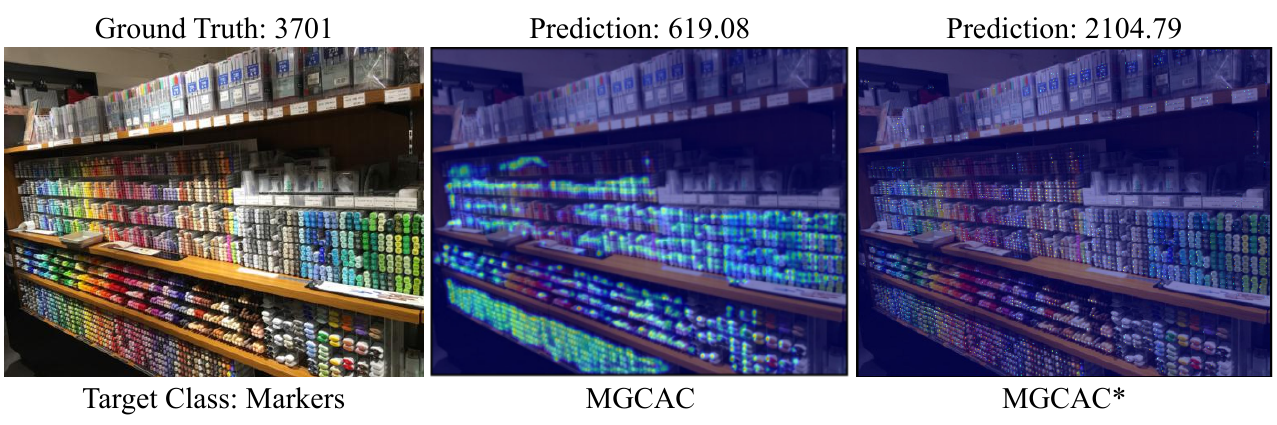} \vspace{-10pt}
\caption{It shows the influence of applying test-time normalization on 1123.jpg in the FSC-147 dataset using our MGCAC, where * indicates applying test-time normalization.}
\label{fig:divide}
\end{figure}

We construct a variant of our MGCAC (denoted as MGCAC$^\ast$) by adopting test-time normalization on MGCAC. Specifically, we utilize the divide-and-conquer technique in~\cite{liu2022countr}, where each query image is cropped by $M \times M$ pieces if the average reference to query area ratio is less than a threshold $T$. For each cropped sub-image, we first interpolate it to the query's original size followed by feeding the resized image into the model. The predicted count of the entire image is the sum of the predicted count of $M^2$ cropped images. We set $M = 8$ and $T = 0.0002$. As shown in Figure~\ref{fig:divide}, MGCAC$^\ast$ exhibits improvement on 1123.jpg (i.e. the outlier with the largest object count from the test dataset) and achieves SOTA on test MAE and test RMSE (cf. Table 2 in the main paper). Given the bias of MAE and RMSE towards extremely large-count query images (cf. Section 3.3 in the main paper), we demonstrate that our model is able to easily hack existing metrics by applying Test-Time Normalization on large-count query images (e.g., 1123.jpg). To provide a more balanced evaluation, we suggest utilizing metrics like NAE or RMSE, which can better reflect the overall performance of the model across the entire dataset.
\begin{figure}[t]
\centering
\includegraphics[width=0.95\linewidth]{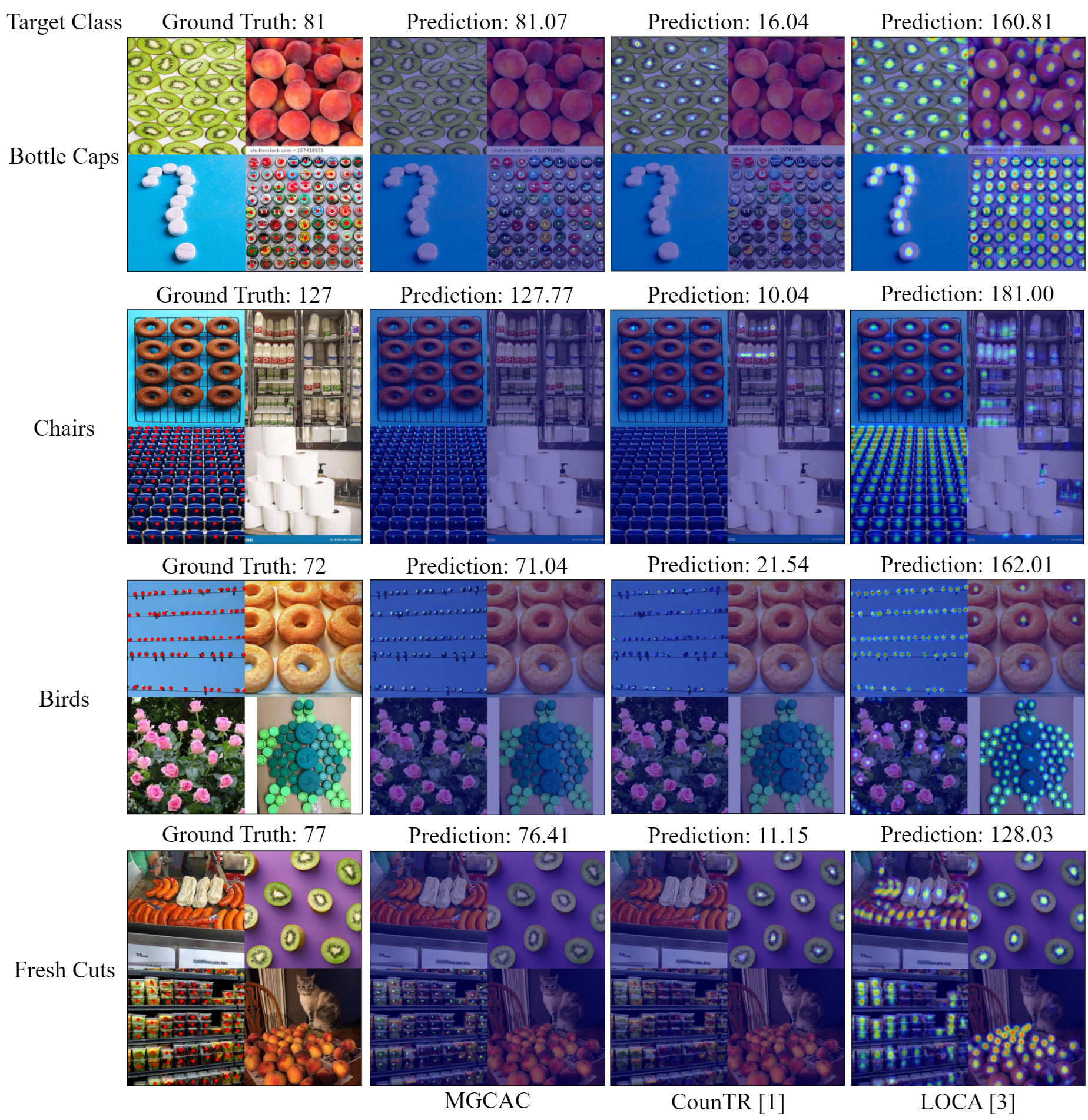}
\vspace{-10pt}
\caption{\textbf{Qualitative Comparisons on Mosaic Dataset based on FSC-147.} The first column indicates the target classes, the second column displays queries with ground truth annotated by red dots and the following columns present the predictions from the respective CAC models (please kindly zoom in to better visualize the highlighted regions of heat/response maps).}
\label{fig:more_results_mosaic}\vspace{-15pt}
\end{figure}
\section{Model Architecture}
\begin{figure}[t]
\centering
\includegraphics[width=\linewidth]{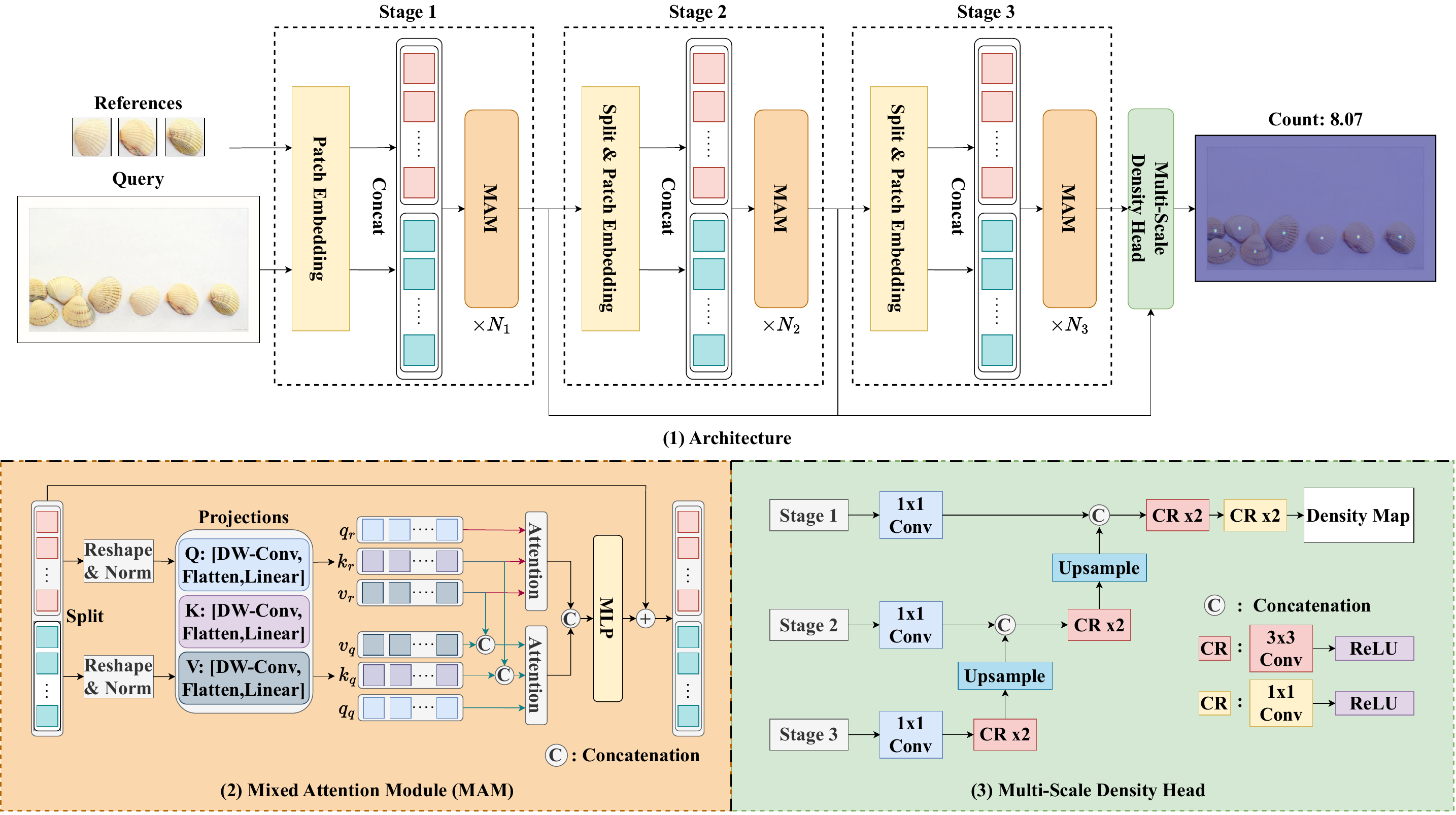}
\vspace{-10pt}
\caption{\textbf{MGCAC Architecture}. We employ a multi-stage extract-and-match module from~\cite{cui2022mixformer} and a density head as an additional strong CAC baseline candidate.}
\label{fig:mgcac}
\end{figure}

In this section, we provide the details of our proposed MGCAC as shown in Figure~\ref{fig:mgcac}, which is another strong baseline candidate for CAC.
Deviating from the standard CAC pipeline that exclusively matches high-level features, the MGCAC model follows MixFormer~\cite{cui2022mixformer} architecture design, which simultaneously extracts visual features from the reference images and augments query features by matching a similarity map among the references and the query, and resulting in finer matching results. We also aggregate features across model stages to capture the nuance of multi-scale features. 
More explanations of specific components are described as follows.
\subsection{Extract-And-Match Module}
To capture reference-specific features and enhance cross-branch interactions, we adopt the MAM (Mixed-Attention Module) from~\cite{cui2022mixformer}. As shown in Figure~\ref{fig:mgcac}(2), both reference and query branches are first performed with convolutional projection to preserve spatial context. Then, the reference feature strengthens its representation through self-attention, and the query branch captures the query’s pixel-wise correlation through cross-attention with reference features. By refining and matching within the same block, the model excels at shaping discriminative features that generate robust cross-attention similarity maps. Unlike previous CAC models that perform feature extraction separately on query and reference and then interact through matching, MAM facilitates frequent cross-branch information exchange.
\subsection{Multi-Scale Enhancement for Matching and Density Map Estimation}
After feature extraction and matching, most of the prior works adopt a simple density head consisting of several upsampling and convolution operations on single-scale feature maps and achieve acceptable performance. To better preserve the information of fine features from early layers/stages across scales, we respectively perform matching on features across stage 1, 2, and 3 of the model, where the feature dimensions are decreased by half as the stage progresses. Then, we utilize a U-Net-like expansion path as the density head to fuse query and similarity-related features across scales. After fusing feature maps, we utilize 1 × 1 convolutions to compress multi-channel feature maps into a single final density map as shown in Figure~\ref{fig:mgcac}(3).
\section{More Evaluation Results}
\subsection{Qualitative Comparisons on Mosaic Dataset}
In Figure~\ref{fig:more_results_mosaic}, we also show qualitative comparisons of our MGCAC and recent SOTA CAC models (e.g. LOCA and CounTR) on the proposed mosaic dataset. LOCA~\cite{djukic2022low} fails to distinguish the corresponding objects and localize 
all objects in the queries. CounTR \cite{liu2022countr} either captures wrong objects (row 1, 2, and 4) or gives inaccurate counts (row 3). In contrast, our MGCAC effectively localizes the right objects as the one provided in the reference and gives accurate count predictions.
The results are consistent with Figure 2 in the main paper, which utilized real image inputs. Therefore, our proposed mosaic dataset can effectively reflect the performance of the model under multi-class scenarios of the real world, enabling a more comprehensive evaluation of CAC models.
\subsection {Cross-Dataset Generalization.} 
A strong CAC model should excel in counting objects on scenarios that differ from the training dataset in terms of perspectives, scales, illumination, etc. To evaluate the ability to generalize on other datasets, we train the proposed method on the FSC-147 dataset and then evaluate on the CARPK testing set without the use of fine-tuning on the CARPK training set. Specifically, since the CARPK dataset mainly consists of car images, we remove the car category from the FSC-147 dataset during the training stage. In Table~\ref{table:cross_data}, our models outperform the non-fine-tuned baselines, which demonstrates MGCAC's generalizability in diverse counting scenarios.
\subsection{Impact of Number of References Images (i.e. $n$)}
We investigate our model performance in relation to the number of reference images, where $n \le 3$. In Table~\ref{tab:n_ref_images}, the results indicate that using our training protocol, our robust model architecture effectively captures reference representative features and improves performance as the number of reference images increases.
\begin{table}[ht]
\centering
\caption {Cross-dataset evaluation results compared with other SOTA models where `fine-tuned' means the model is further fine-tuned using the CARPK training data.}
\begin{tabular}{p{3cm} c c c}
\toprule
Model & Fine-tuned  & MAE & RMSE \\  \hline
BMNet \cite{shi2022represent} & \checkmark & 8.05 & 9.7 \\
BMNet+ \cite{shi2022represent} & \checkmark & 5.76 & 7.83 \\
CounTR \cite{liu2022countr} &\checkmark & \textbf{5.75} & \textbf{7.45} \\ \hline
BMNet \cite{shi2022represent}& $\times$ & 14.61 & 24.60\\
BMNet+ \cite{shi2022represent}& $\times$ & 10.44 & 13.77\\
LOCA \cite{djukic2022low} & $\times$ & 9.97 & 12.51 \\
\textbf{MGCAC (Ours)} & $\times$ & 
\textbf{6.41} & \textbf{8.81} \\ \bottomrule
\end{tabular}\label{table:cross_data}
\end{table}
\begin{table}[ht]
\centering
\caption{Evaluation results of MGCAC using different numbers (i.e. $n$) of reference images on FSC-147.}
\label{tab:n_ref_images}
\centering
\resizebox{0.75\columnwidth}{!}{%
\setlength{\tabcolsep}{5pt}
\begin{tabular}{ccccccccc} \toprule
\multirow{2}{*}{$n$} & \multicolumn{4}{c}{VAL}     & \multicolumn{4}{c}{TEST}     \\ 
                   & MAE   & RMSE  & NAE  & SRE  & MAE   & RMSE   & NAE  & SRE  \\ \hline
1                  & 17.78 & 68.9  & 0.19 & 3.05 & 14.55 & 111.04 & 0.16 & 2.79 \\
2                  & 12.80 & 58.83 & 0.13 & 2.34 & 11.03 & 107.09 & \textbf{0.15} & \textbf{4.24} \\
3                  & \textbf{11.00} & \textbf{51.42} & \textbf{0.12} & \textbf{2.05} & \textbf{10.46} & \textbf{96.60}  & 0.16 & 6.17 \\ \bottomrule
\end{tabular}
}
\end{table}
\begin{figure}[ht]
\centering \vspace{-10pt}
\includegraphics[width=\columnwidth]{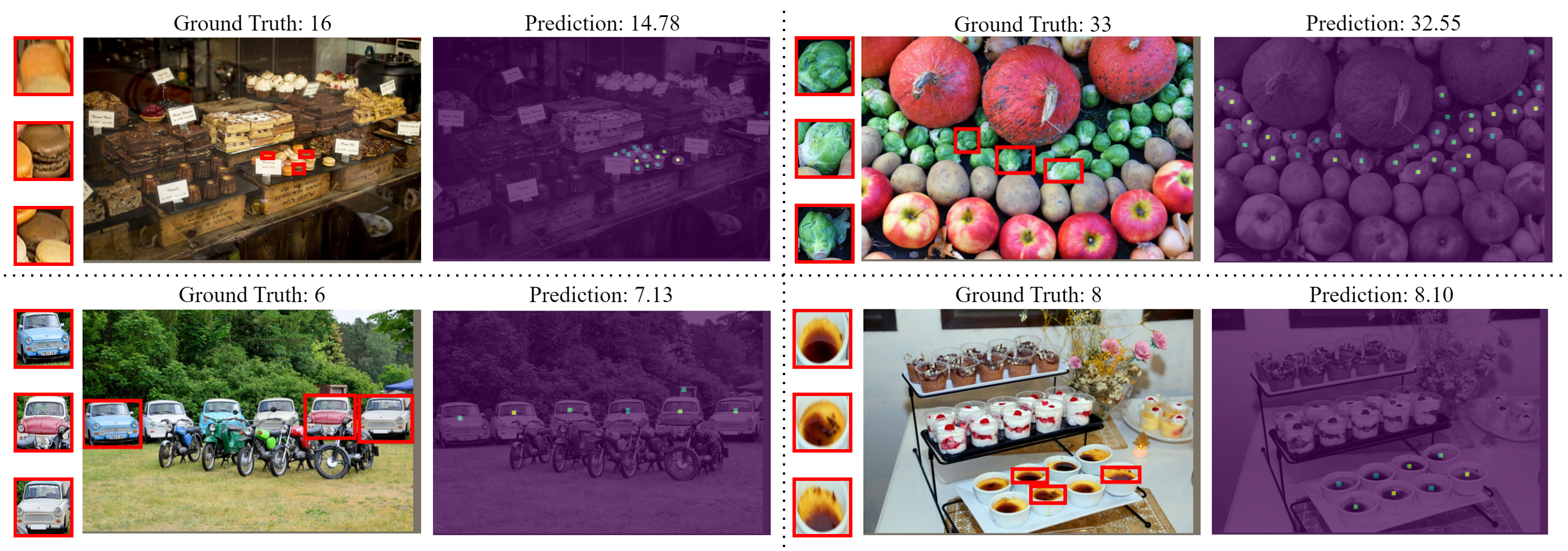} \vspace{-10pt}
\caption{\textbf{More real-world examples of MGCAC.} The references are annotated by bounding boxes.}
\label{fig:more_results_real} 
\end{figure}
%
%
\FloatBarrier
\bibliographystyle{splncs04}
\bibliography{references/cac, references/loss, references/tracking}